\def\year{2018}\relax
\documentclass[letterpaper]{article} 
\usepackage{aaai18}  
\usepackage{times}  
\usepackage{helvet}  
\usepackage{courier}  
\usepackage{url}  
\usepackage{graphicx}  

\usepackage{amsmath}
\usepackage{amssymb}
\usepackage{mathrsfs}
\usepackage{array}
\usepackage{multirow}
\usepackage{graphicx}
\usepackage{epstopdf}
\usepackage[list=true]{subcaption}

\begin{document}
%
\title{Adversarial Network Embedding}
\author{Quanyu Dai$^{1}$, Qiang Li$^{1,2}$, Jian Tang$^{3, 4}$, Dan Wang$^{1}$\\
$^{1}$Department of Computing, The Hong Kong Polytechnic University, Hong Kong\\
$^{2}$School of Software, FEIT, The University of Technology Sydney, Australia\\
$^{3}$HEC Montreal, Canada\\
$^{4}$Montreal Institute for Learning Algorithms, Canada\\
csqydai@comp.polyu.edu.hk, leetsiang.cloud@gmail.com, tangjianpku@gmail.com, csdwang@comp.polyu.edu.hk}
\maketitle

\begin{abstract}
 Learning low-dimensional representations of networks has proved effective in a variety of tasks such as node classification, link prediction and network visualization. Existing methods can effectively encode different structural properties into the representations, such as neighborhood connectivity patterns, global structural role similarities and other high-order proximities. However, except for objectives to capture network structural properties, most of them suffer from lack of additional constraints for enhancing the robustness of representations. In this paper, we aim to exploit the strengths of generative adversarial networks in capturing latent features, and investigate its contribution in learning stable and robust graph representations. Specifically, we propose an Adversarial Network Embedding (ANE) framework, which leverages the adversarial learning principle to regularize the representation learning. It consists of two components, i.e., a structure preserving component and an adversarial learning component. The former component aims to capture network structural properties, while the latter contributes to learning robust representations by matching the posterior distribution of the latent representations to given priors. As shown by the empirical results, our method is competitive with or superior to state-of-the-art approaches on benchmark network embedding tasks.
\end{abstract}

\section{Introduction}
 Graph is a natural way of organizing data objects with complicated relationships, and encodes rich information of nodes in the graph.
 For example, paper citation networks capture the information of innovation flow, and can reflect topic relatedness between papers. To analyze graphs, an efficient and effective way is to learn low-dimensional representations for nodes in the graph, i.e., node embedding~\cite{KDD-14-Bryan,WWW-15-Jian,CIKM-15-SsCao}. The learned representations should encode meaningful semantic, relational and structural information, so that they can be used as features for downstream tasks such as network visualization, link prediction and node classification. Network embedding is a challenging research problem because of the high-dimensionality, sparsity and non-linearity of the graph data.

 \begin{figure}[t]
    \centering
    \subcaptionbox{IDW.}{
        \includegraphics[angle=0, width=0.22\textwidth]{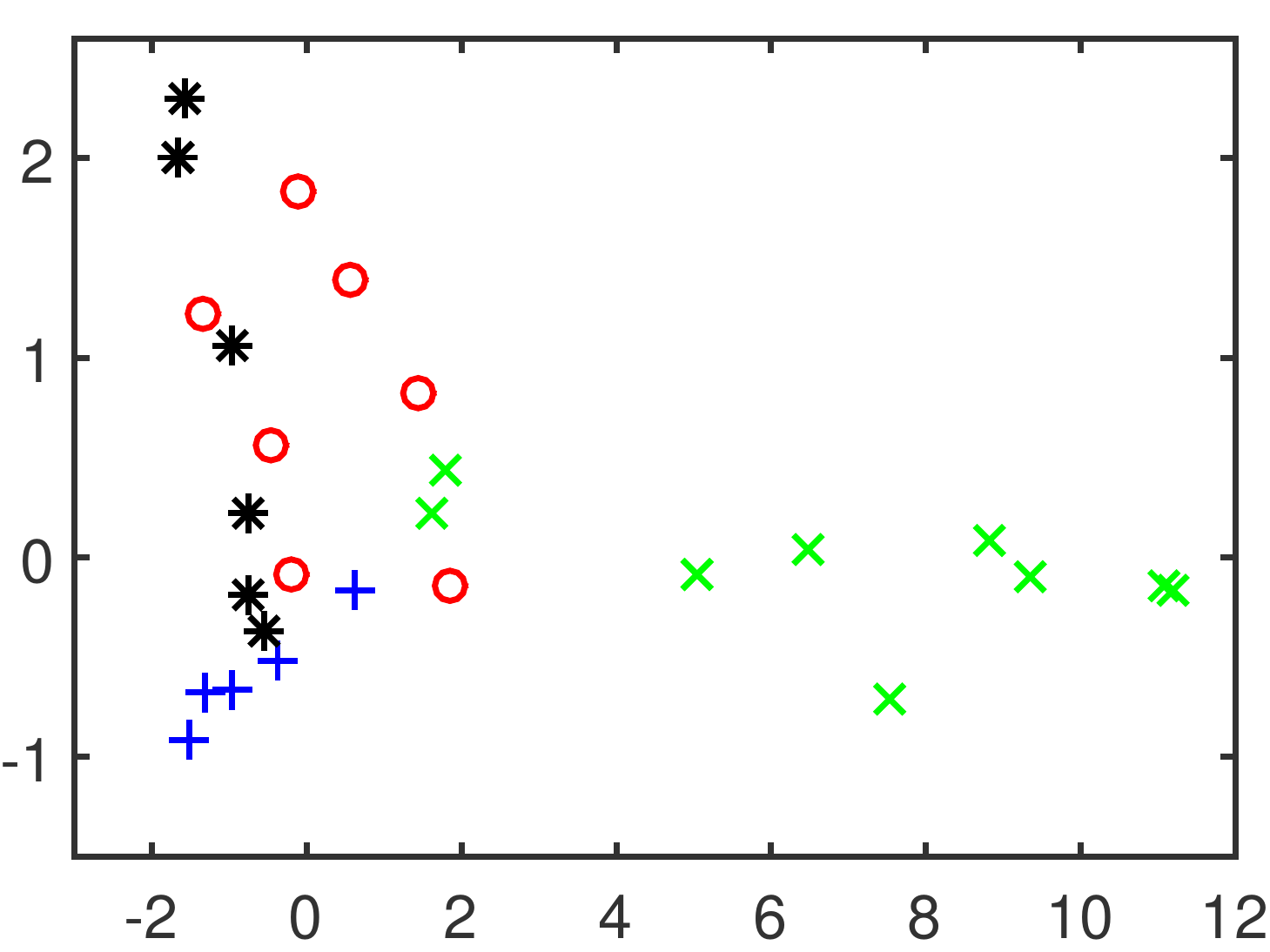}
        \label{fig:IDW-karate}}
    \subcaptionbox{AIDW.}{
        \includegraphics[angle=0, width=0.22\textwidth]{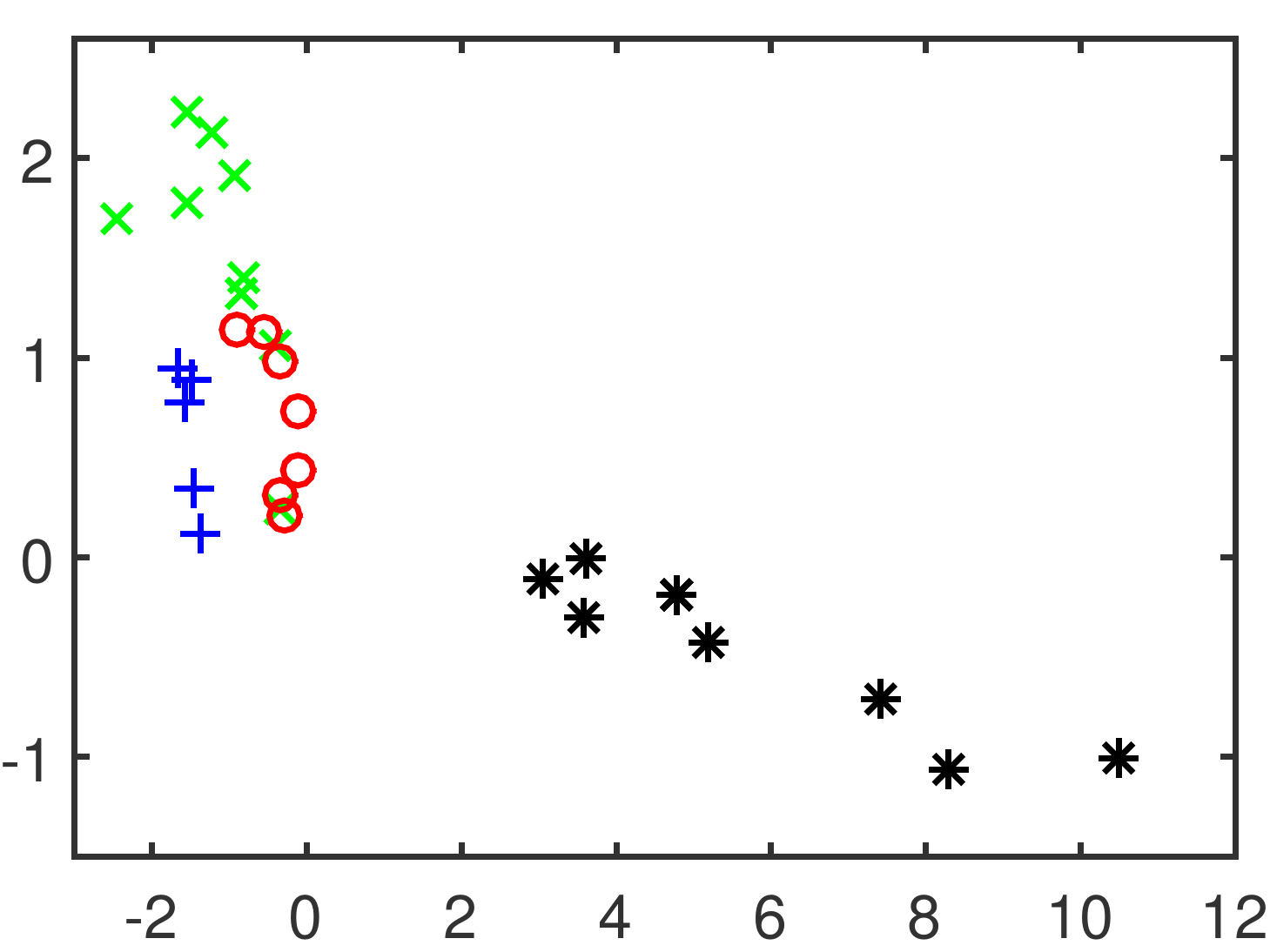}
        \label{fig:AIDW-karate}}
    \caption{Visualization of two dimensional representations of Zachary's Karate network~\cite{JAR-77-Zachary} from Inductive DeepWalk (IDW) and Adversarial Inductive DeepWalk (AIDW). Different colors represent different communities detected by modularity-based clustering. As shown by the figure, AIDW can better capture community structure information, demonstrating that adversarial learning contributes to learning more meaningful and robust representations.}
    \label{karate}
 \end{figure}

 In recent years, many methods for network embedding have been proposed, such as DeepWalk~\cite{KDD-14-Bryan}, LINE~\cite{WWW-15-Jian} and node2vec~\cite{KDD-16-Grover}. They aim to capture various connectivity patterns in network during representation learning. These patterns include relations of local neighborhood connectivity, first and second order proximities, global structural role similarities (i.e. structural equivalence), and other high-order proximities. As demonstrated in the literature, network embedding methods were shown to be more effective in many network analysis tasks than some classical approaches, such as Common Neighbors~\cite{JASIS-Liben-NowellK07} and Spectral Clustering~\cite{DMKD-11-TangL}.

 Though existing methods are effective in structure preserving with different carefully designed objectives, they suffer from lack of additional constraints for enhancing the robustness of the learned representations. When processing noisy network data, which is very common in real-world applications, these unsupervised network embedding techniques can easily result in poor representations. Thus, it is critical to consider some amount of uncertainty in the process of representation learning. One famous technique for robust representation learning in unsupervised manner is denoising autoencoder~\cite{JMLR-VincentLLBM10}. It obtains stable and robust representations by recovering clean input from the corrupted one, that is denoising criterion. In~\cite{AAAI-16-SsCao}, the authors have applied this criterion for network embedding. Recently, many generative adversarial models~\cite{ICLR-16-RadfordMC,ICLR-16-MakhzaniSJG,ICLR-17-Donahue,ICLR-17-Vincent} have also been proposed for learning robust and reusable representations. They have been shown to be effective in learning representations for image~\cite{ICLR-16-RadfordMC} and text data~\cite{NIPS-workshop-16-Glover}. However, none of such models have been specially designed for dealing with graph data.

 In this paper, we propose a novel approach called Adversarial Network Embedding (ANE) for learning robust network representations by leveraging the principle of adversarial learning~\cite{NIPS-14-GoodfellowPMXWOCB}. In addition to optimize the objective for preserving structure, a process of adversarial learning is introduced for modeling the data uncertainty. Figure~\ref{karate} presents an illustrative example with the well-known Zachary's Karate network on the effect of adversarial learning. By comparing representations of two schemes without/with adversarial learning regularization, it can be easily found that the latter scheme obtains more meaningful and robust representations.

 More specifically, ANE naturally combines a structure preserving component and an adversarial learning component in a unified framework. The former component can help capture network structural properties, while the latter contributes to the learning of more robust representations through adversarial training with samples from some prior distribution. For structure preserving, we propose an inductive variant of DeepWalk that is suitable for our ANE framework. It maintains random walk for exploring neighborhoods of nodes and optimizes similar objective function, but employs parameterized function to generate embedding vectors. Besides, the adversarial learning component consists of two parts, i.e., a generator and a discriminator. It is acting as a regularizer for learning stable and robust feature extractor, which is achieved by imposing a prior distribution on the embedding vectors through adversarial training. To the best of our knowledge, this is the first work to design network embedding model with the adversarial learning principle. We empirically evaluate the proposed ANE approach through network visualization and node classification on benchmark datasets. The qualitative and quantitative results prove the effectiveness of our method.

\section{Related Work}

\subsection{Network Embedding Methods}
 In recent years, many unsupervised network embedding methods have been proposed, which can be divided into three groups according to the techniques they use, i.e., probabilistic methods, matrix factorization based methods and autoencoder based methods. The probabilistic methods include DeepWalk~\cite{KDD-14-Bryan}, LINE~\cite{WWW-15-Jian}, node2vec~\cite{KDD-16-Grover} and so on. DeepWalk firstly obtains node sequences from the original graph through random walk, and then learns the latent representations using Skip-gram model~\cite{NIPS-13-Tomas} by regarding node sequences as word sentences. LINE tries to preserve first-order and second-order proximities in two separate objective functions, and then directly concatenates the representations. In~\cite{KDD-16-Grover}, the authors proposed to use biased random walk to determine neighboring structure, which can strike a balance between homophily and structural equivalence. It is actually a variant of DeepWalk.

 Matrix factorization based methods first preprocess the adjacency matrix to capture different kinds of high-order proximities and then decompose the processed matrix to obtain graph embeddings. For example, GraRep~\cite{CIKM-15-SsCao} employs positive pointwise mutual information (PPMI) matrix as the preprocessing based on a proof of the equivalence between a $k$-step random walk in DeepWalk and a $k$-step probability transition matrix. HOPE~\cite{KDD-16-Mingdong} preprocesses the adjacency matrix of the directed graph with high-order proximity measurements, such as Katz Index~\cite{Katz-Index-1953}, which can help capture asymmetric transitivity property. M-NMF~\cite{AAAI-17-XiaoW} learns embeddings that can well capture community structure by building upon the modularity based community detection model~\cite{PhysRevE-06-Newman}.

 Autoencoder is a widely used model for learning compact representations of high-dimensional data, which aims to preserve as much information in the latent space as possible for the reconstruction of the original data~\cite{Sci-06-Hinton}. DNGR~\cite{AAAI-16-SsCao} firstly calculates the PPMI matrix, and then learns the representations through stacked denosing autoencoder. SDNE~\cite{KDD-16-DxW} is a variant of stacked autoencoder which adds a constraint in the loss function to force the connected nodes to have similar embedding vectors. In~\cite{NIPS-16-Kipf}, the authors proposed a variational graph autoencoder (VGAE) by using a graph convolutional network~\cite{ICLR-16-KipfW} encoder for capturing network structural properties. Compared to variational autoencoder (VAE)~\cite{ICLR-13-KingmaW}, our ANE approach explicitly regularizes the posterior distribution of the latent space while VAE only assumes a prior distribution.

\subsection{Generative Adversarial Networks}
 Generative Adversarial Networks (GANs)~\cite{NIPS-14-GoodfellowPMXWOCB} are deep generative models, of which the framework consists of two components, i.e., a generator and a discriminator. GANs can be formulated as a minimax adversarial game, where the generator aims to map data samples from some prior distribution to data space, while the discriminator tries to tell fake samples from real data. This framework is not directly suitable for unsupervised representation learning, due to the lack of explicit structure for inference.

 There are three possible solutions for this problem as demonstrated by existing works. Firstly, some works managed to integrate some structures  into the framework to do inference, i.e., projecting sample in data space back into the space of latent features, such as BiGAN~\cite{ICLR-17-Donahue}, ALI~\cite{ICLR-17-Vincent} and EBGAN~\cite{ICLR-17-ZhaoML}. These methods can learn robust representations in many applications, such as image classification~\cite{ICLR-17-Donahue} and document retrieval~\cite{NIPS-workshop-16-Glover}. The second approach is to generate representations from the hidden layer of the discriminator, like DCGANs~\cite{ICLR-16-RadfordMC}. By employing fractionally-strided convolutional layers, DCGANs can learn expressive image representations from both the generator and discriminator networks for supervised tasks. The third idea is to use adversarial learning process to regularize the representations. One successful practice is the Adversarial Autoencoders~\cite{ICLR-16-MakhzaniSJG}, which can learn powerful representations from unlabeled data without any supervision.

\section{Adversarial Network Embedding}
 In this section, we will first introduce the problem definition and notations to be used. Then, we will present an overview of the proposed adversarial network embedding framework, followed by detailed descriptions of each component.

\subsection{Problem Definition and Notations}
 Network embedding is aimed at learning meaningful representations for nodes in information network. An information network can be denoted as $\mathcal{G}=(V, E, A)$, where $V$ is the node set, $E$ is a set of edges with each representing the relationship between a pair of nodes, and $A$ is a weighted adjacency matrix with its entries quantifying the strength of the corresponding relations. Particularly, the value of each entry in $A$ is either 0 or 1 in an unweighted graph specifying whether an edge exists between two nodes. Given an information network $\mathcal{G}$, network embedding is doing a mapping from nodes $v_i \in V$ to low-dimensional vectors $\boldsymbol{u_i} \in R^d$ with the formal format as follows: $f: V \mapsto U$, where $\boldsymbol{u_i}^T$ is the $i$th row of $U$ ($U \in R^{N \times d}$, $N=|V|$) and $d$ is the dimension of representations. We call $U$ representation matrix. These representations should encode structural information of networks.

\subsection{An Overview of the Framework}
\begin{figure}[t]
    \centering
    \includegraphics[width=0.95\columnwidth]{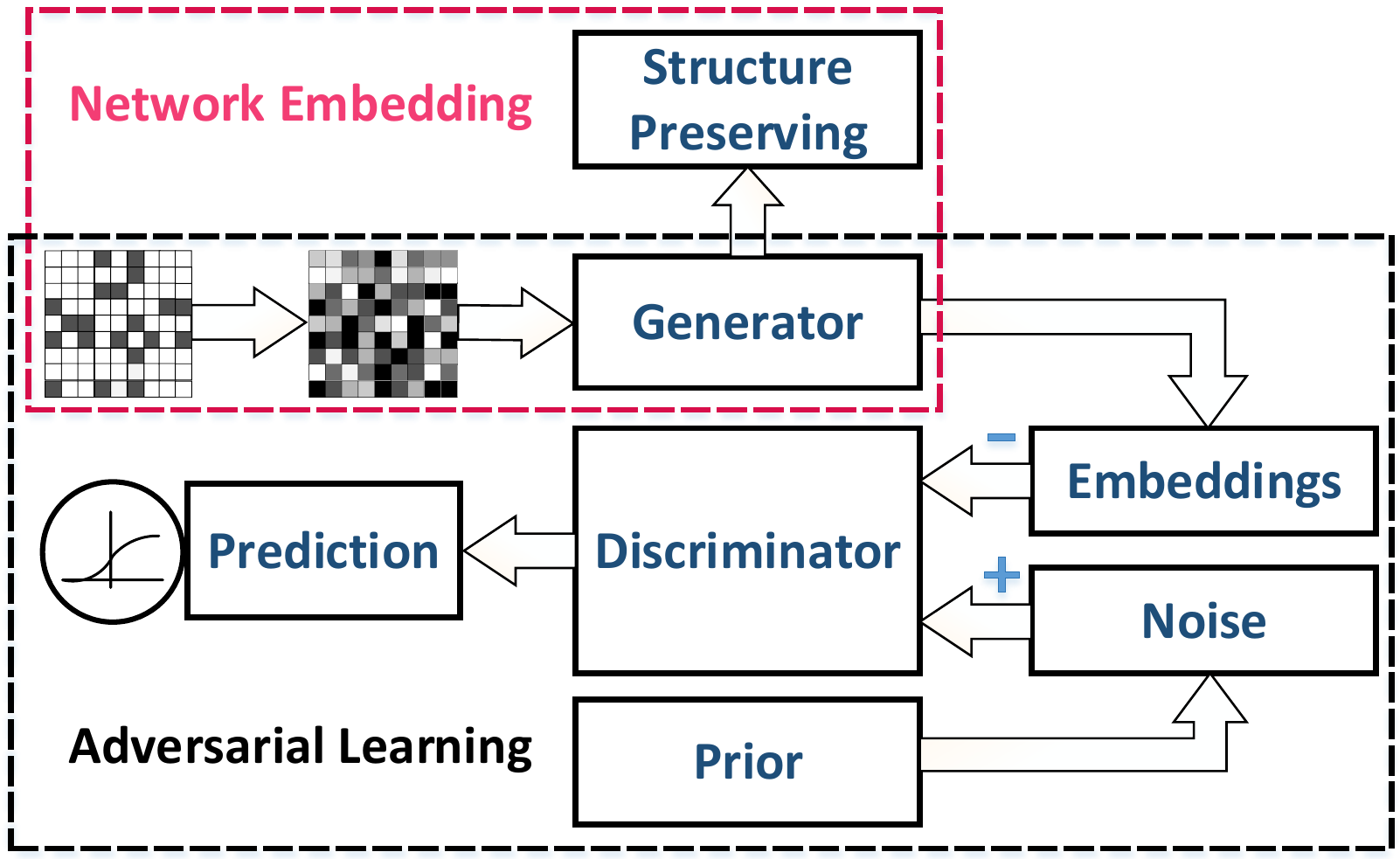}
    \caption{Adversarial Network Embedding Framework}
    \label{ANE-Framework}
 \end{figure}

 In this work, we leverage adversarial learning principle to help learn stable and robust representations. Figure~\ref{ANE-Framework} shows the proposed framework of~\textit{Adversarial Network Embedding} (ANE), which mainly consists of two components, i.e., a structure preserving component and an adversarial learning component. Specifically, the structure preserving component is dedicated to encoding network structural information into the representations. These information include the local neighborhood connectivity patterns, global structural role similarities, and other high-order proximities. There are many possible alternatives for the implementation of this component. Actually, existing methods~\cite{KDD-14-Bryan,WWW-15-Jian,CIKM-15-SsCao} can be considered as structure preserving models, but without any constraints to help enhance the robustness of the representations. In this paper, we propose an inductive DeepWalk for structure preserving. It maintains random walk for exploring neighborhoods of nodes and optimizes similar objective function, but employs parameterized function $G(\cdot)$ to generate embedding vectors. In the training process, parameters of $G(\cdot)$ are directly updated instead of the embedding vectors. Besides, the adversarial learning component consists of two parts, i.e., a generator $G(\cdot)$ and a discriminator $D(\cdot)$. It is acting as a regularizer for learning stable and robust feature extractor, which is achieved by imposing a prior distribution on the embedding vectors through adversarial training. It needs to emphasize that the parameterized function $G(\cdot)$ is shared by both the structure preserving component and the adversarial learning component. These two components will update the parameters of $G(\cdot)$ alternatively in the training process.

\subsection{Graph Preprocessing}
 In real-world applications, information network is usually extremely sparse, which may result in serious over-fitting problem when training deep models. To help alleviate the sparsity problem, one commonly used method is to preprocess the adjacency matrix with high-order proximities~\cite{WWW-15-Jian,CIKM-15-SsCao}. In this paper, we employ the shifted PPMI matrix $X$~\cite{NIPS-14-LevyG} as input features for the generator\footnote{Note that we can use other ways to preprocess raw graph data to obtain the input feature X with lower dimension for large graphs. One simple way is to directly use existing scalable methods, e.g. DeepWalk and LINE, to obtain initial embeddings X as input.}, which is defined as
 \begin{equation}
  X_{ij} = \max\{\log(\frac{M_{ij}}{\sum_{k}M_{kj}})-\log(\beta), 0\},
 \end{equation}
 where $M=\hat A+\hat A^2+\cdots +\hat A^t$ can capture different high-order proximities, $\hat{A}$ is the $1$-step probability transition matrix obtained from the weighted adjacency matrix $A$ after a row-wise normalization, and $\beta$ is set to $\frac{1}{N}$ in this paper. Row vector $\boldsymbol{x_i}^T$ in $X$ is the feature vector characterizing the context information of node $v_i$ in the graph $\mathcal{G}$, but with high-dimension.

\subsection{Structure Preserving Model}
 Ideally, existing unsupervised network embedding methods can be utilized as structure preserving component in our framework for encoding node dependencies into representations. However, many of them are transductive methods with an embedding lookup as embedding generator such as DeepWalk and LINE, which are not directly suitable for the generator of the adversarial learning component since we utilize parameterized generator as standard GANs. With parameterized generator, our framework can well deal with networks with node attributes and explore nonlinear properties of network with deep learning models. In this work, we design an inductive variant of DeepWalk that is applicable for both weighted and unweighted graphs. Theoretically, it can also generalize to unseen nodes for networks with node attributes as some inductive methods do~\cite{ICML-16-YangCS,NIPS-17-HamiltonYL}, but we do not explore it in this paper. Besides, we also investigate to use denoising autoencoder~\cite{JMLR-VincentLLBM10} as the structure preserving component.

\subsubsection{Inductive DeepWalk (IDW)}
 The IDW model uses random walk to sample node sequences as that in DeepWalk. Starting from each node $v_i$, $\eta$ sequences are randomly sampled with the length as $l$. In every step, a new node is randomly selected from the neighbors of the current node with the probability proportional to the corresponding weight in matrix $A$. To improve efficiency, the alias table method~\cite{KDD-LiARS14} is employed to sample node from the candidate node set in every sampling step. It only takes $O(1)$ time in a single sampling step. Then, positive node pairs can be constructed from node sequences. For every node sequence $\mathcal{W}$, we determine the positive target-context pairs as the set $\{(w_i,w_j): |i-j|<s\}$, where $w_i$ is the $i$th node in sequence $\mathcal{W}$ and $s$ denotes the context size.

 Similar to Skip-gram~\cite{NIPS-13-Tomas}, a node $v_i$ has two different representations, i.e., a target representation $\boldsymbol{u_i}$ and a context representation $\boldsymbol{u^{\prime}_i}$, which are generated by the target generator $G(\cdot; \boldsymbol{\theta_1})$ and context generator $F(\cdot;\boldsymbol{\theta^{\prime}_1})$, respectively. The generators are parameterized functions which are implemented with neural networks in this work. Given row vector $\boldsymbol{x_i}^T$ in $X$ corresponding to node $v_i$, we have $\boldsymbol{u_i} = G(\boldsymbol{x_i};\boldsymbol{\theta_1})$ and $\boldsymbol{u^{\prime}_i} = F(\boldsymbol{x_i};\boldsymbol{\theta^{\prime}_1})$. To capture network structural properties, we define the following objective function for each positive target-context pair $(v_i, v_j)$ with negative sampling approach:
 \begin{equation}\label{IDW-Loss}
 \begin{array}{ll}
 \mathcal{O}_{IDW}(\boldsymbol{\theta_1};\boldsymbol{\theta^{\prime}_1}) = \log \sigma(F(\boldsymbol{x_j};\boldsymbol{\theta^{\prime}_1})^{T} G(\boldsymbol{x_i};\boldsymbol{\theta_1})) + \\
 \sum_{n=1}^{K}\mathbb{E}_{v_n\sim P_n(v)}[\log \sigma(-F(\boldsymbol{x_n};\boldsymbol{\theta^{\prime}_1})^{T} G(\boldsymbol{x_i};\boldsymbol{\theta_1}))],
 \end{array}
 \end{equation}
 where $\sigma(x)=1/(1+exp(-x))$ is the sigmoid function, $K$ is the number of negative samples for each positive pair, $P_n(v)$ is the noise distribution for sampling negative context nodes, $\boldsymbol{\theta_1}$ and $\boldsymbol{\theta^{\prime}_1}$ are parameters to be learnt. As suggested in~\cite{NIPS-13-Tomas}, $P_n(v) = d_v^{3/4}/\sum_{v_i\in V}d_{v_i}^{3/4}$ can achieve quite good performance in practice, where $d_v$ is the degree of node $v$.

\subsection{Adversarial Learning}
 The adversarial learning component is employed to regularize the representations. It consists of a generator $G(\cdot;\boldsymbol{\theta_1})$ and a discriminator $D(\cdot;\boldsymbol{\theta_2})$. Specifically, $G(\cdot;\boldsymbol{\theta_1})$ represents a non-linear transformation of input high-dimensional features to embedding vectors. $D(\cdot;\boldsymbol{\theta_2})$ represents the probability of a sample coming from real data. The generator function is shared with the structure preserving component. Different from GANs~\cite{NIPS-14-GoodfellowPMXWOCB}, in our framework, a prior distribution $p(\boldsymbol{z})$ is selected as the data distribution for generating real data, while the embedding vectors are regarded as fake samples. In the training process, the discriminator is trained to tell apart the prior samples from the embedding vectors, while the generator is aimed to fit embedding vectors to the prior distribution. This process can be considered as a two-player minimax game with the generator and discriminator playing against each other. The utility function of the discriminator is:
 \begin{align}\label{ANE-Discriminator}
 \begin{split}
 \mathcal{O}_D(\boldsymbol{\theta_2})=\;&\mathbb{E}_{\boldsymbol{z}\sim p(\boldsymbol{z})}[\log D(\boldsymbol{z};\boldsymbol{\theta_2})] + \\
 &\mathbb{E}_{\boldsymbol{x}}[\log(1-D(G(\boldsymbol{x};\boldsymbol{\theta_1});\boldsymbol{\theta_2}))].
 \end{split}
 \end{align}
 In order to camouflage its output as prior samples, the generator is trained to improve the following payoff:
 \begin{equation}\label{ANE-Generator}
 \mathcal{O}_G(\boldsymbol{\theta_1})=\mathbb{E}_{\boldsymbol{x}}[\log(D(G(\boldsymbol{x};\boldsymbol{\theta_1});\boldsymbol{\theta_2}))].
 \end{equation}

 We argue that the adversarial learning component can help improve the learned representations in terms of robustness and structural meanings. We instantiate our framework with two structure preserving models, i.e., inductive DeepWalk (IDW) and denoising autoencoder (DAE). We call the ANE framework with IDW as Adversarial Inductive DeepWalk (AIDW) for easy illustration. Actually, with DAE as the structure preserving component, the ANE framework will become an adversarial autoencoder~\cite{ICLR-16-MakhzaniSJG}, and we represent it as ADAE to highlight the importance of the denoising criterion in learning representations.

 During adversarial learning, it is also important to choose a proper prior distribution. Like many practices in GANs research~\cite{ICLR-16-RadfordMC,ICLR-16-MakhzaniSJG,ICLR-17-Donahue}, the prior distribution is usually defined as Uniform or Gaussian noise which enables GANs to learn meaningful and robust representations against uncertainty. In our experiments, we also considered the ANE framework with both kinds of prior distributions but find no significant difference. One possible reason is both kinds of uncertainty can help the ANE framework to achieve a certain level of robustness against noise. It is likely that a careful choice of prior distribution, possibly guided by prior domain knowledge, may further improve application-specific performance.

\subsection{Algorithm}
 To implement the ANE approach, we consider a joint training procedure with two phases, including a structure preserving phase and an adversarial learning phase. In the structure preserving phase, we optimize objective function~(\ref{IDW-Loss}) for AIDW. In the adversarial learning phase, a prior distribution is imposed on representations through a minimax optimization problem. Firstly, the discriminator is trained to distinguish between prior samples and embedding vectors. Then, the parameters of the generator are updated to fit the embedding vectors to prior space to fool the discriminator. Besides, some tricks proposed in~\cite{ICML-ArjovskyCB17} can be employed to help improve the stability of learning and avoid the mode collapse problem in traditional GANs training.

\section{Experiments}
 \begin{figure*}[t]
    \centering
    \subcaptionbox{DeepWalk.}{
        \includegraphics[angle=0, width=0.19\textwidth]{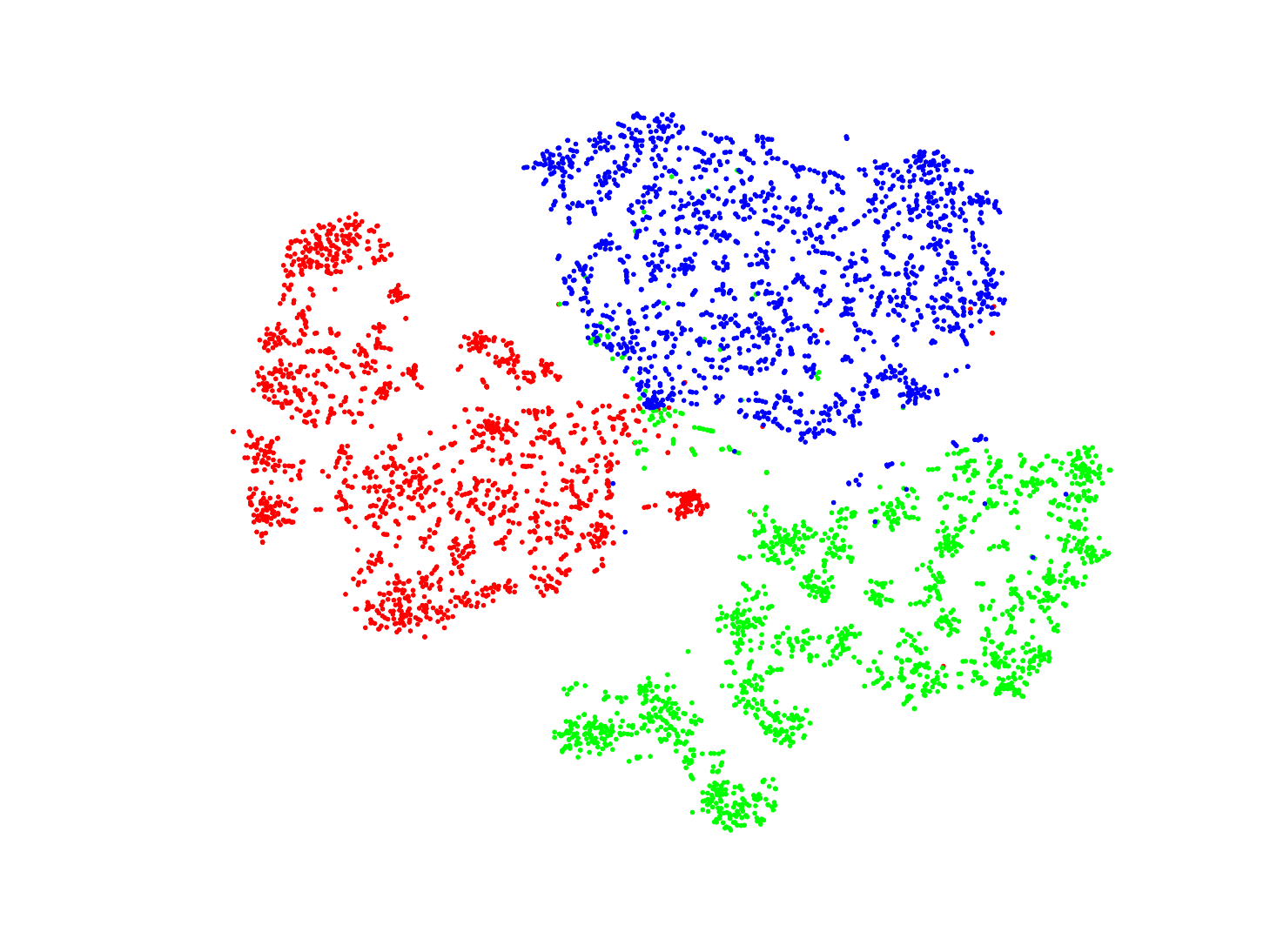}
        \label{fig:DW-SCI}}
    \hspace{-1em}
    \subcaptionbox{LINE.}{
        \includegraphics[angle=0, width=0.19\textwidth]{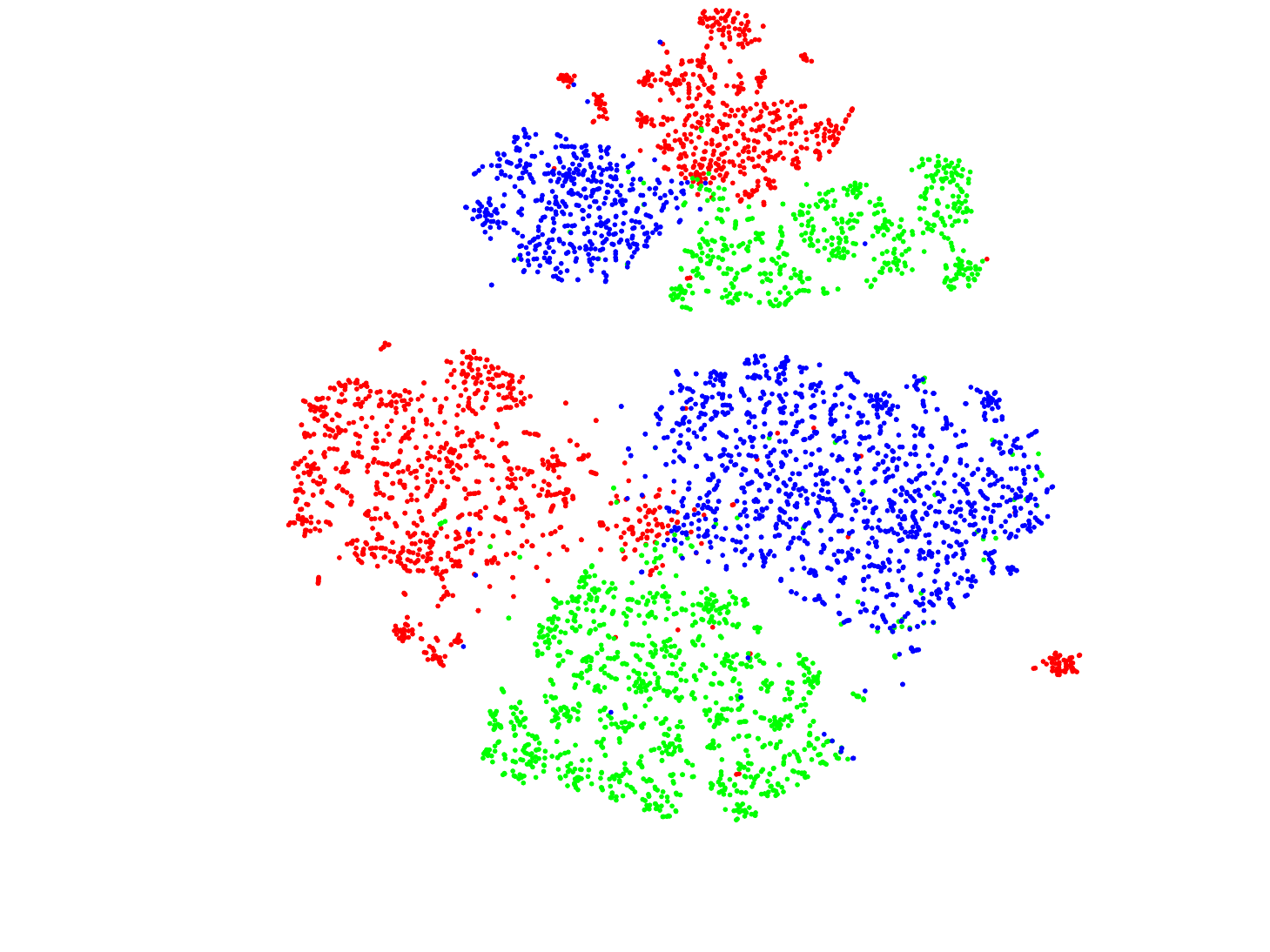}
        \label{fig:LINE-SCI}}
    \hspace{-1em}
    \subcaptionbox{node2vec.}{
        \includegraphics[angle=0, width=0.19\textwidth]{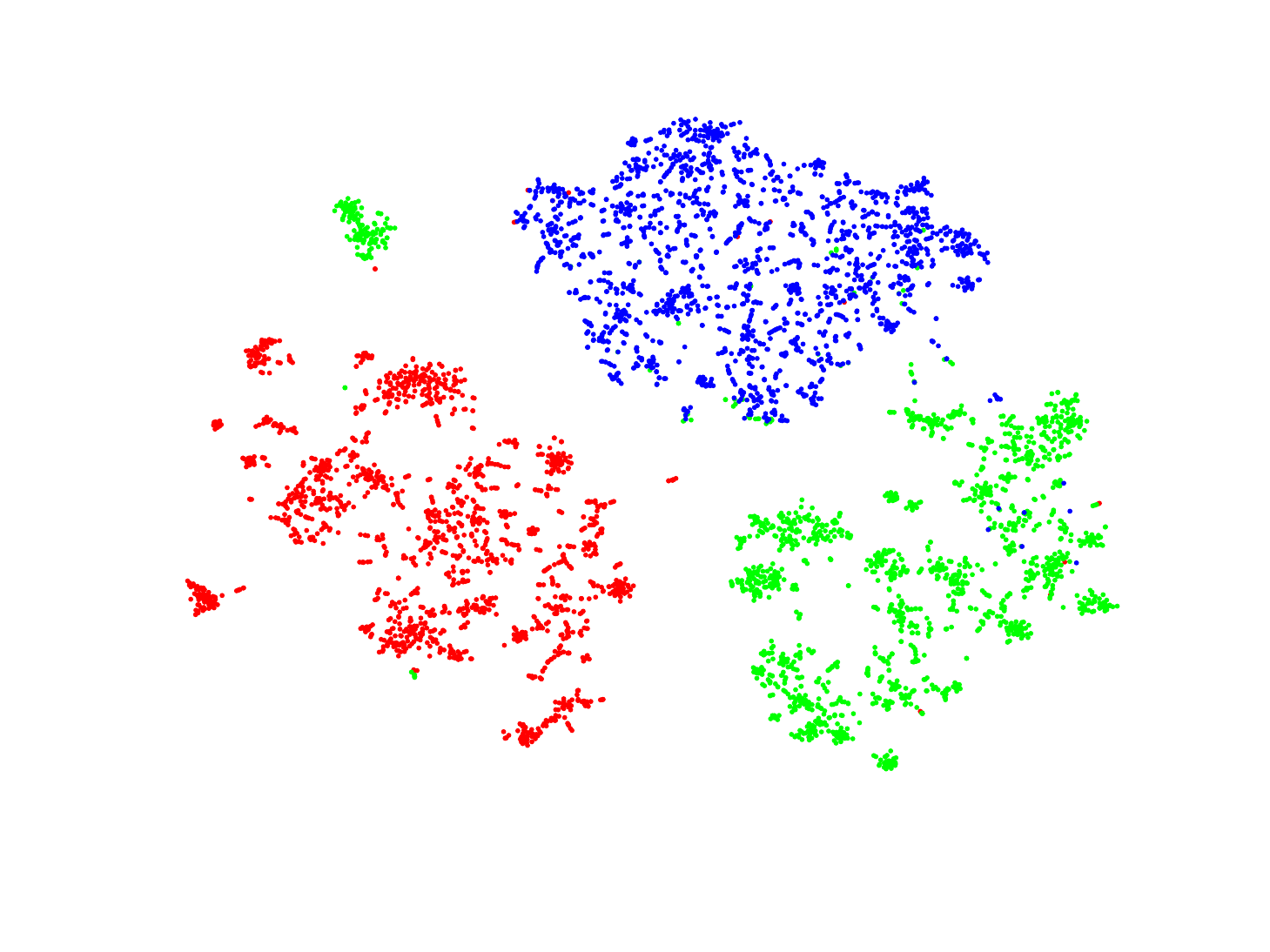}
        \label{fig:node2vec-SCI}}
    \hspace{-1em}
    \subcaptionbox{IDW.}{
        \includegraphics[angle=0, width=0.19\textwidth]{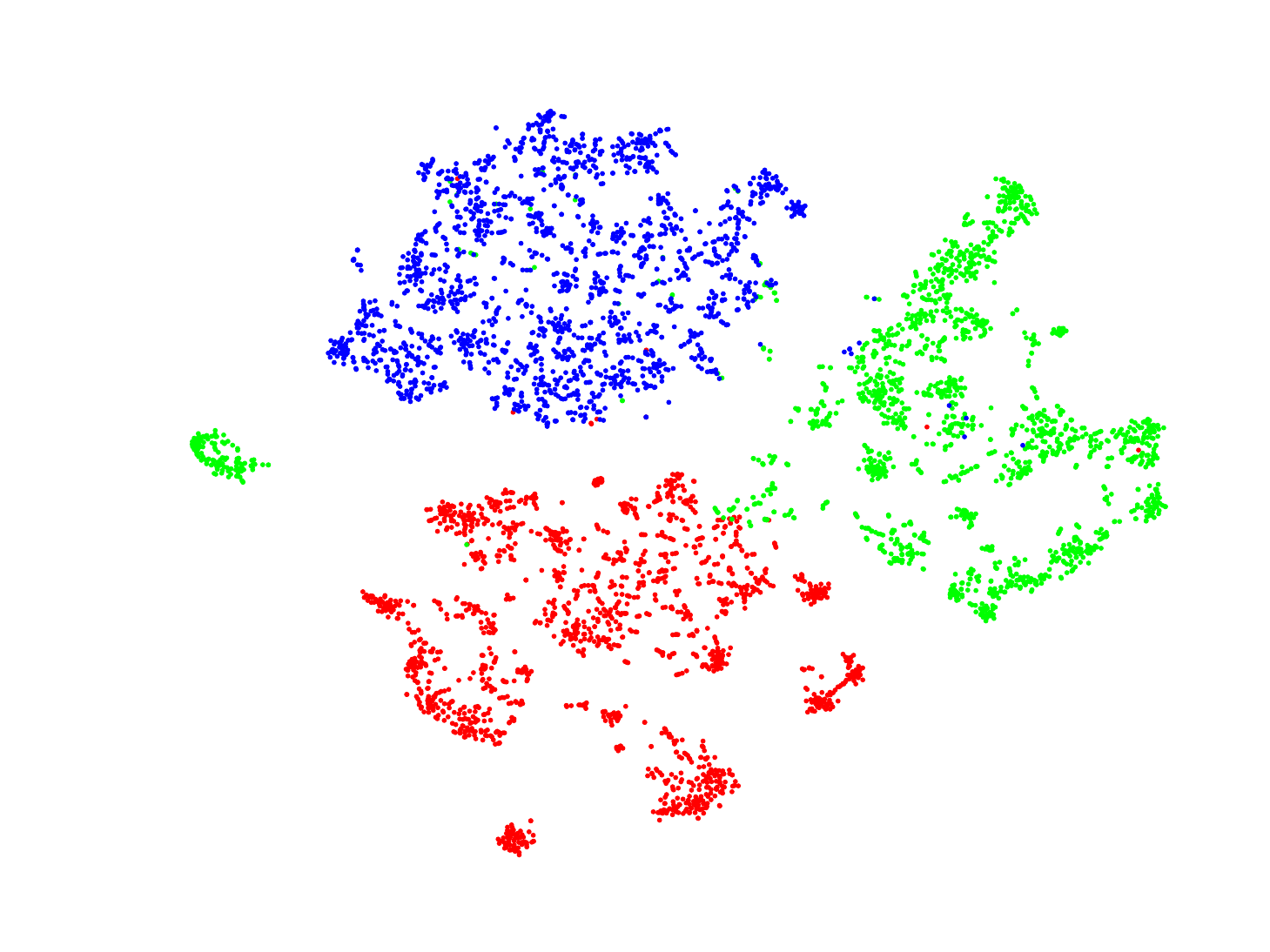}
        \label{fig:IDW-SCI}}
    \hspace{-1em}
    \subcaptionbox{AIDW.}{
        \includegraphics[angle=0, width=0.19\textwidth]{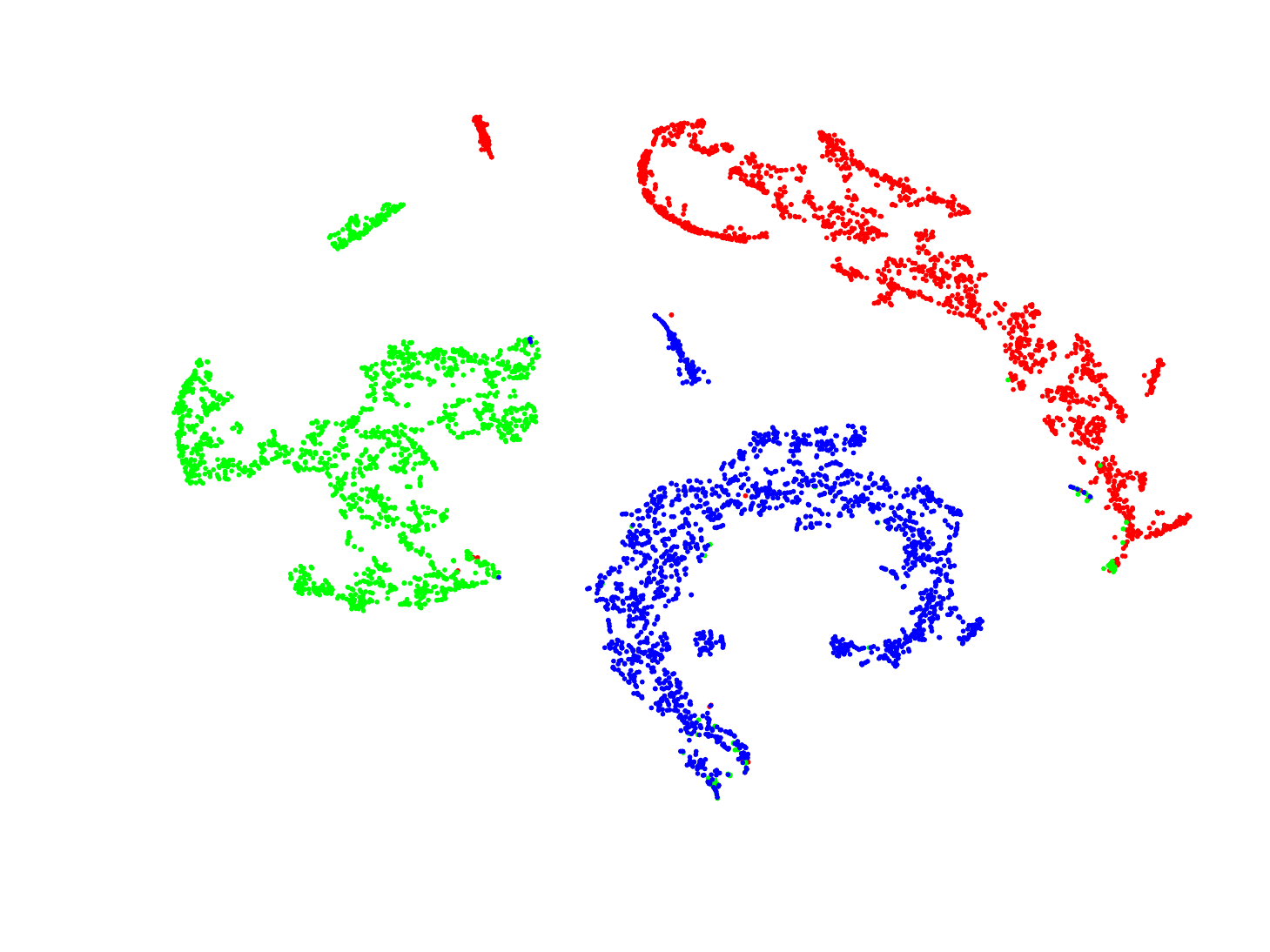}
        \label{fig:AIDW-SCI}}
    \caption{Visualization of Cit-DBLP dataset. Each point represents one paper. Different colors correspond to different publication divisions. Red: ``Information Science'', blue: ``ACM Transactions on Graphics'', green: ``Human-Computer Interaction''. }
    \label{visualization}
 \end{figure*}

\subsection{Experiment Settings}
\subsubsection{Datasets}
 We conduct experiments on four real-world datasets with the statistics presented in Table~\ref{tab-dataset}, where $\mathcal{C}$ denotes the label set. Cora and Citeseer are paper citation networks constructed by~\cite{Retr-00-McCallumNRS}. Wiki~\cite{AI-M-08-Sen} is a network with nodes as web pages and edges as the hyperlinks between web pages. We regard these three networks as undirected networks, and do some preprocessing on the original datasets by deleting self-loops and nodes with zero degree. Cit-DBLP is a paper citation network extracted from DBLP dataset~\cite{KDD-08-TangZYLZS}.

 \begin{table}[t]\small
     \centering
     \caption{Statistics of datasets}
     \begin{tabular}{ l | c | c | c }
      \hline
      \hline			
      Dataset & $\mid V\mid$ & $\mid E\mid$ & $\mid\mathcal{C}\mid$ \\
      Cora & 2,708 & 5,278 & 7 \\
      Citeseer & 3,264 & 4,551 & 6 \\
      Wiki & 2,363 & 11,596 & 17 \\
      Cit-DBLP & 5,318 & 28,085 & 3 \\
      \hline
      \hline
     \end{tabular}
     \label{tab-dataset}
 \end{table}

\subsubsection{Baselines}
 We compare our model with several baseline methods, including DeepWalk, LINE, GraRep and node2vec. There are many other network embedding methods, but we do not consider them here, because their performances are inferior to these baseline models as shown in corresponding papers. The descriptions of the baselines are as follows.

 \begin{itemize}
    \item \textbf{DeepWalk}~\cite{KDD-14-Bryan}: DeepWalk first transforms the network into node sequences by truncated random walk, and then uses it as input to the Skip-gram model to learn representations.
    \item \textbf{LINE}~\cite{WWW-15-Jian}: LINE can preserve both first-order and second-order proximities for undirected graph through modeling node co-occurrence probability and node conditional probability.
    \item \textbf{GraRep}~\cite{CIKM-15-SsCao}: GraRep preserves node proximities by constructing different $k$-step probability transition matrices.
    \item \textbf{node2vec}~\cite{KDD-16-Grover}: node2vec develops a biased random walk procedure to explore neighborhood of a node, which can strike a balance between local properties and global properties of a network.
 \end{itemize}

 Besides, we consider inductive DeepWalk and denoising autoencoder as another two baseline methods. Note that both of them employ shifted PPMI matrix as preprocessing.

 \subsubsection{Parameter Settings}
 For LINE, we follow the settings of parameters in~\cite{WWW-15-Jian}. The embedding vectors are normalized by L2-norm. Besides, we specially preprocess the original sparse networks by adding two-hop neighbors to low degree nodes. For both DeepWalk and node2vec, the window size $s$, the walk length $l$ and the number of walks $\eta$ per node are set to 10, 80 and 10, respectively, for fair comparison. For GraRep, the maximum matrix transition step is set to 4, and the settings of other parameters follow those in~\cite{CIKM-15-SsCao}. Note that the dimension of representations for all methods are set to 128 for fair comparison.

 For our methods, we only use the most simple structure for the generator. Specifically, the generator is a single-layer network with leaky ReLU activations (with a leak of 0.2) and batch normalization~\cite{ICML-IoffeS15} on the output. The shifted PPMI matrix $X$ is obtained by setting $t$ as 4 for Cora and Citeseer, and 3 for Wiki. For inductive DeepWalk, the number of negative samples $K$ is set to 5, and other parameters are set the same as DeepWalk. For denoising autoencoder, it has only one hidden layer with dimension as 128. For the discriminator of the framework, it is a three-layer neural networks, with the layer structure as 512-512-1. For the first two layers, we use leaky ReLU activations (with leak of 0.2) and batch normalization. For the output layer, we use sigmoid activation. For AIDW and ADAE, the settings of the structure preserving component are the same as those of IDW and DAE, respectively. The prior distribution of the adversarial learning component is set to $z_i\sim U[-1,1]$. We use RMSProp optimizer with learning rate as 0.001.

\subsection{Network Visualization}

 \begin{table*}[t]\small
 \centering
 \caption{Accuracy (\%) of multi-class classification on Cora}
 \begin{tabular}{ c | c | c | c | c | c | c | c | c | c }
  \hline
  \hline
  \%Labeled Nodes & 10\% & 20\% & 30\% & 40\% & 50\% & 60\% & 70\% & 80\% & 90\% \\
   DeepWalk & 71.43 & 73.83 & 75.61 & 76.92 & 77.79 & 77.78 & 78.47 & 79.17 & 79.04 \\
   LINE & 71.26 & 74.50 & 76.04 & 76.81 & 77.68 & 77.99 & 78.46 & 79.00 & 79.11 \\
   GraRep & 74.78 & 76.78 & 78.56 & 78.99 & 79.39 & 79.85 & 79.96 & 80.94 & 81.29 \\
   node2vec & 75.06 & 78.49 & 80.06 & 80.94 & 81.52 & 82.07 & 82.39 & 83.28 & 83.17 \\
   \hline
   DAE & 75.21 & 78.07 & 79.39 & 80.51 & 80.91 & 81.41 & 82.36 & 83.23 & 83.32 \\
   ADAE & 75.01 & 77.45 & 79.65 & 80.96 & 81.64 & 82.11 & 82.62 & 83.52 & 84.10 \\
   \hline
   IDW & 66.32 & 72.21 & 75.23 & 76.65 & 77.66 & 78.32 & 79.20 & 79.93 & 80.63 \\
   AIDW & \textbf{76.93} & \textbf{79.50} & \textbf{81.31} & \textbf{82.01} & \textbf{82.28} & \textbf{83.03} & \textbf{83.23} & \textbf{84.46} & \textbf{84.21} \\
  \hline
  \hline
 \end{tabular}
 \label{tab-cora}
 \end{table*}

 \begin{table*}[!htb]\small
 \centering
 \caption{Accuracy (\%) of multi-class classification on Citeseer}
 \begin{tabular}{ c | c | c | c | c | c | c | c | c | c }
  \hline
  \hline
  \%Labeled Nodes & 10\% & 20\% & 30\% & 40\% & 50\% & 60\% & 70\% & 80\% & 90\% \\	
   DeepWalk & 49.45 & 52.68 & 54.60 & 55.71 & 56.44 & 57.04 & 57.42 & 58.04 & 59.11 \\
   LINE & 47.70 & 51.04 & 52.95 & 54.19 & 55.00 & 55.82 & 56.02 & 57.08 & 57.52 \\
   GraRep & 51.62 & 53.29 & 53.59 & 53.83 & 54.55 & 54.62 & 54.97 & 54.90 & 56.27 \\
   node2vec & 52.50 & 55.47 & 56.66 & 57.70 & 58.81 & 59.26 & 60.10 & 60.34 & 60.58 \\
   \hline
   DAE & 51.08 & 54.54 & 55.84 & 56.50 & 57.47 & 58.30 & 58.50 & 59.19 & 60.15 \\
   ADAE & 52.40 & 55.58 & 56.64 & 57.32 & 58.34 & 59.60 & 60.27 & 60.63 & 61.31 \\
   \hline
   IDW & 45.45 & 50.47 & 52.32 & 53.38 & 54.75 & 55.18 & 55.98 & 56.23 & 57.19 \\
   AIDW & \textbf{53.25} & \textbf{56.76} & \textbf{57.95} & \textbf{59.06} & \textbf{59.45} & \textbf{59.95} & \textbf{60.28} & \textbf{60.87} & \textbf{62.26} \\
  \hline
  \hline
 \end{tabular}
 \label{tab-citeseer}
 \end{table*}

 \begin{table*}[!htb]\small
 \centering
 \caption{Accuracy (\%) of multi-class classification on Wiki}
 \begin{tabular}{ c | c | c | c | c | c | c | c | c | c }
  \hline
  \hline
  \%Labeled Nodes & 10\% & 20\% & 30\% & 40\% & 50\% & 60\% & 70\% & 80\% & 90\% \\
   DeepWalk & 57.23 & 61.22 & 63.53 & 64.68 & 65.96 & 66.24 & 67.17 & 68.63 & 68.69 \\
   LINE & 56.29 & 61.63 & 63.98 & 65.43 & 66.25 & 67.04 & 67.94 & 68.86 & 68.61 \\
   GraRep & \textbf{57.68} & 61.14 & 62.73 & 63.89 & 64.86 & 65.55 & 66.01 & 67.55 & 68.02 \\
   node2vec & 57.61 & 61.52 & 63.47 & 64.83 & 65.54 & 66.16 & 67.17 & 68.44 & 68.69 \\
   \hline
   DAE & 57.08 & 61.63 & 63.71 & 65.20 & 66.84 & 67.41 & 67.91 & 69.03 & 69.45 \\
   ADAE & 57.24 & 61.67 & 63.85 & 65.34 & 66.67 & 67.11 & 67.79 & 69.68 & 70.59 \\
   \hline
   IDW & 56.01 & 60.77 & 63.08 & 64.37 & 65.66 & 66.47 & 67.15 & 67.86 & 68.52 \\
   AIDW & 57.43 & \textbf{62.14} & \textbf{64.18} & \textbf{65.53} & \textbf{67.07} & \textbf{68.00} & \textbf{69.44} & \textbf{71.63} & \textbf{72.03} \\
  \hline
  \hline
 \end{tabular}
 \label{tab-wiki}
 \end{table*}

 Network visualization is an indispensable way to analyze high-dimensional graph data, which can help reveal intrinsic structure of the data intuitively~\cite{WWW-TangLZM16}. In this section, we visualize the representations of nodes generated by several different models using~\textit{t-SNE}~\cite{JMLR-08-Maaten}. We construct a paper citation network, namely Cit-DBLP, from DBLP with papers from three different publication divisions, including Information Sciences, ACM Transactions on Graphics and Human-Computer Interaction. Some statistics of this dataset have been presented in Table~\ref{tab-dataset}. These papers are naturally classified into three categories based on the research fields they belong to.

 Figure~\ref{visualization} shows the visualization of embedding vectors obtained from different models using~\textit{t-SNE} tookit under the same parameter configuration. For both DeepWalk and LINE, papers from different categories are mixed with each other in the center of the figure. For LINE, there are 6 clusters with each category corresponding to two separate clusters, which is in conflict with the true structure of the network. Besides, the boundaries between different clusters are not clear. The visualizations of node2vec and IDW form three main clusters, which are better than those of DeepWalk and LINE. However, the boundary between blue cluster and green cluster for node2vec is not clear, while that of red cluster and green cluster is a little messy for IDW. AIDW performs better compared with baseline methods. We can observe that the visualization of AIDW has three clusters with quite large margin between each other. Furthermore, each cluster is linearly separable with another cluster in the figure, which can not be achieved by other baselines as showed by the figures. Intuitively, this experiment demonstrates that adversarial learning regularization can help learn more meaningful and robust representations.

\subsection{Node Classification}

 The label information can indicate interests, beliefs or other characteristics of nodes, which can help facilitate many applications, such as friend recommendation in online social networks and targeted advertising. However, in many real-world contexts, only a subset of nodes are labeled. Thus, node classification can be conducted to dig out information of unlabeled nodes. In this section, we conduct multi-class classification on three benchmark datasets, i.e., Cora, Citeseer and Wiki. We range the training ratio from 10\% to 90\% for comprehensive evaluation. All experiments are carried out with support vector classifier in Liblinear package\footnote{https://www.csie.ntu.edu.tw/~cjlin/liblinear/}~\cite{JRML-08-FanCHWL}.

 \begin{figure*}[t]
    \centering
    \subcaptionbox{Dimension.}{
        \includegraphics[angle=0, width=0.29\textwidth]{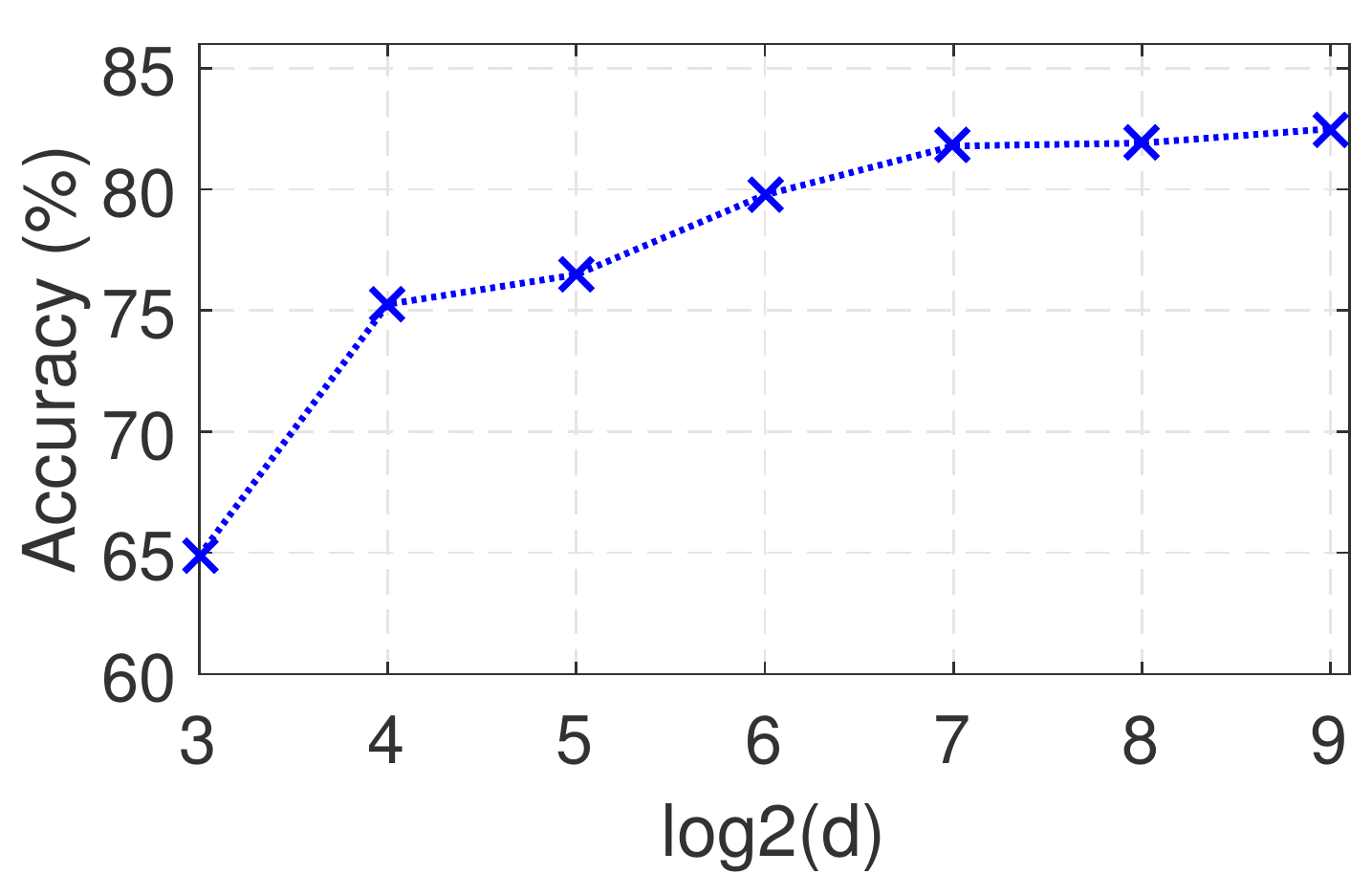}
        \label{fig:dimension}}
    \subcaptionbox{Walk-length.}{
        \includegraphics[angle=0, width=0.29\textwidth]{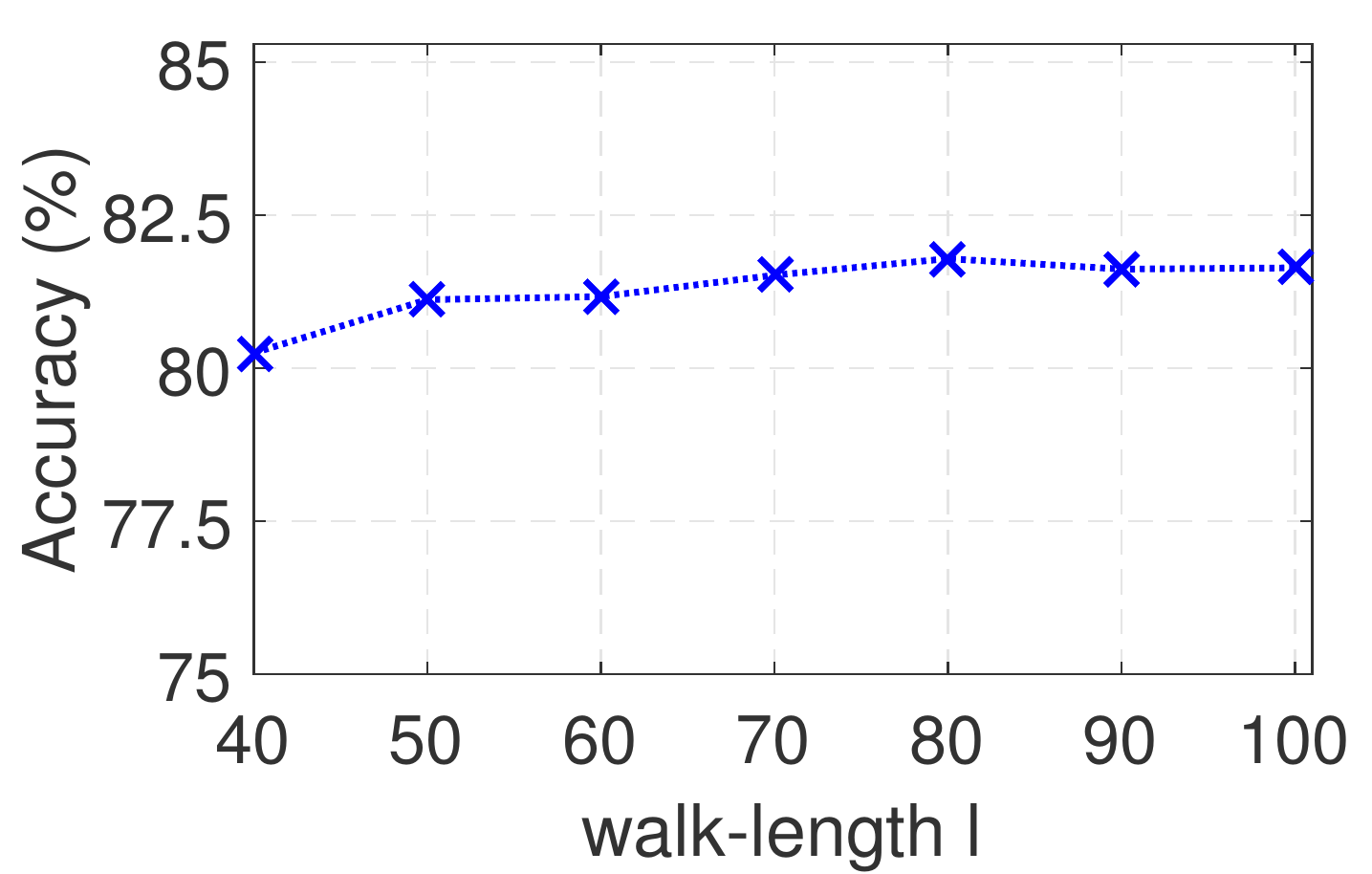}
        \label{fig:walk-length}}
    \subcaptionbox{Context-size.}{
        \includegraphics[angle=0, width=0.29\textwidth]{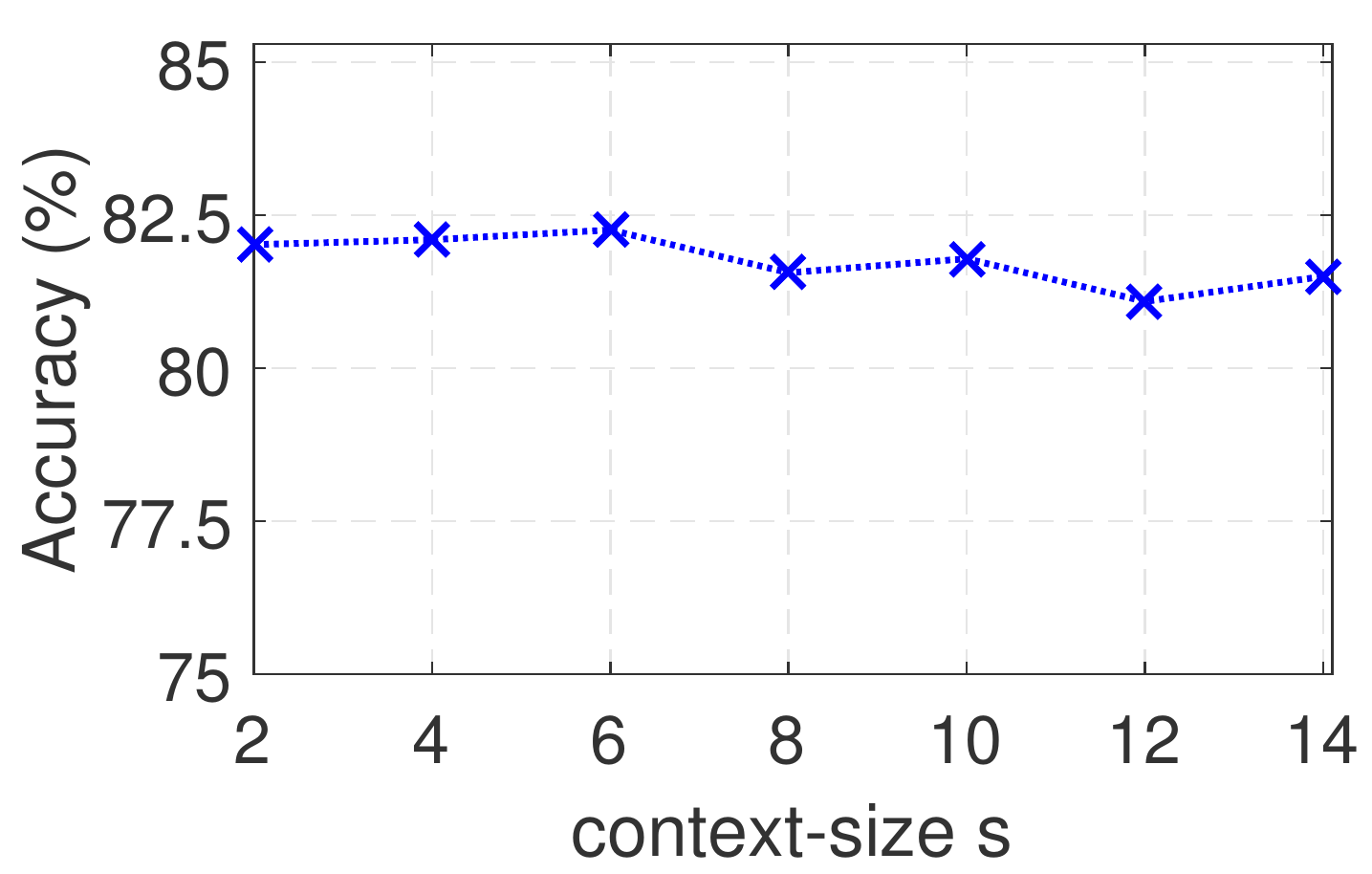}
        \label{fig:context-size}}
    \caption{Parameter sensitivity analysis of AIDW using multi-class classification on Cora with train ratio as 50\%.}
    \label{parameter-sensitivity}
 \end{figure*}

 \textbf{Results and discussion.} To ensure the reliability, we obtain the experimental results by taking an average of that of 10 runs, which are shown in Tables~\ref{tab-cora},~\ref{tab-citeseer} and~\ref{tab-wiki}. We have the following observations:
 \begin{itemize}
   \item IDW produces similar results on Cora and Wiki with DeepWalk, and slightly inferior performance on Citeseer. The proposed model AIDW is built upon IDW with additional adversarial learning component. It consistently outperforms both IDW and DeepWalk on three datasets across all training ratios. For example, on Cora, AIDW gives more than 4\% gain in accuracy over DeepWalk under all training ratio settings. It demonstrates that adversarial learning regularization can significantly improve the robustness and discrimination of the learned representations. The quantitative results also verify our previous qualitative findings in network visualization analysis.

   \item ADAE achieves about 1\% gain in accuracy over DAE on Citeseer when varying the training ratio from 10\% to 90\%, slightly better results on Cora, and comparable performance on Wiki. It shows that ANE framework can also guide the learning of more robust embeddings when building upon DAE. However, we notice that ADAE does not achieve obvious improvements over the corresponding structure preserving model as AIDW does. One reason is that denoising criterion already contributes to learning stable and robust representations~\cite{JMLR-VincentLLBM10}.

   \item Overall, the proposed method AIDW consistently outperforms all the baselines. As shown in Tables~\ref{tab-cora},~\ref{tab-citeseer} and~\ref{tab-wiki}, node2vec produces better results than DeepWalk, LINE and GraRep on average. Our method can further achieve improvements over node2vec. More specifically, AIDW achieves the best classification accuracy on all three benchmark datasets across different training ratios, with only one exception on Wiki with training ratio as 10\%.

 \end{itemize}

\subsection{Model Sensitivity}

 In this section, we investigate the performance of AIDW w.r.t parameters and the type of prior on Cora dataset. Specifically, for parameter sensitivity analysis, we examine how the representation dimension $d$, walk-length $l$ and context-size $s$ affect the performance of node classification with the training ratio as 50\%. Note that except for the parameter being tested, all other parameters are set to default values. We also compare the performance of AIDW with two different priors, i.e., a Gaussian distribution ($\mathcal{N}(0,1)$) and a Uniform distribution ($U[-1, 1]$).

 \begin{figure}[htb!]
    \centering
    \includegraphics[width=0.85\columnwidth]{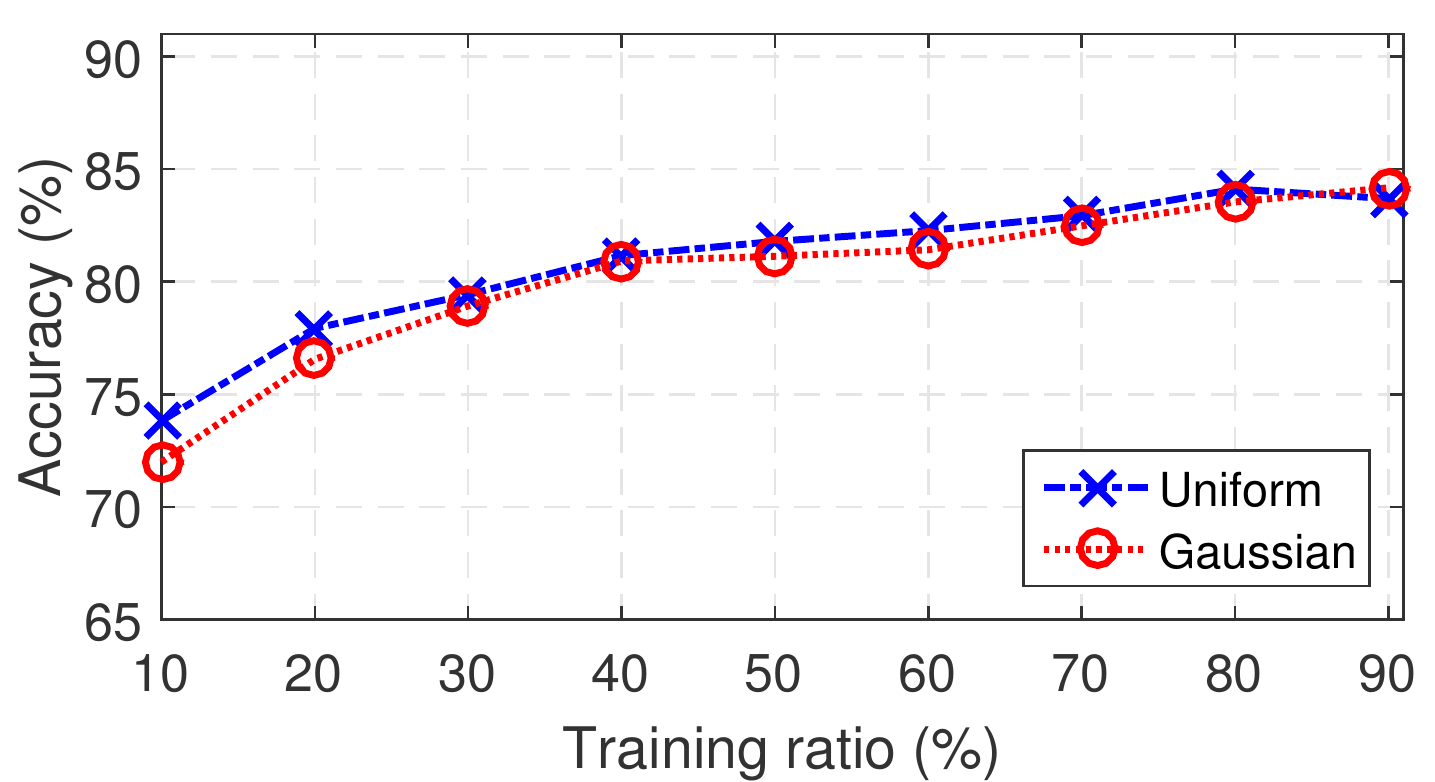}
    \caption{Multi-class classification on Cora with two different priors, i.e., Uniform and Gaussian, on AIDW model.}
    \label{fig:prior}
 \end{figure}

 Figure~\ref{parameter-sensitivity}(a) displays the results  on the test of dimension $d$. When the dimension increases from 8 to 512, the accuracy shows apparent increase at first, and then tends to saturate once the dimension reaches around 128. Besides, the performance of AIDW is not sensitive on walk-length and context-size. As shown in Figure~\ref{parameter-sensitivity}(b), the accuracy slightly increases first, and then becomes stable when the walk-length varies from 40 to 100. With the increase of context-size, the performance keeps stable first, and then slightly degrades after the context-size is over 6, as shown in Figure~\ref{parameter-sensitivity}(c). The degradation might be caused by the noisy neighborhood information brought in by the large context-size, since the average node degree of Cora is just about 1.95.

 Figure~\ref{fig:prior} shows the results of multi-class classification on Cora with training ratio ranging from 10\% to 90\%. The accuracy curve of AIDW with uniform prior is almost coincided with that of AIDW with gaussian prior. It demonstrates that both types of prior can contribute to learning robust representations with no significant difference.

\section{Conclusion}
 An adversarial network embedding framework has been proposed for learning robust graph representations. This framework consists of a structure preserving component and an adversarial learning component. For structure preserving, we proposed inductive DeepWalk to capture network structural properties. For adversarial learning, we formulated a minimax optimization problem to impose a prior distribution on representations to enhance the robustness. Empirical evaluations in network visualization and node classification confirmed the effectiveness of the proposed method.

\section{Acknowledgments}
 The authors would like to thank Dr. Liang Zhang of Data Science Lab at JD.com and Prof. Xiaoming Wu of The Hong Kong Polytechnic University for their valuable discussion. Dan Wang's work is supported in part by HK PolyU G-YBAG.

\bibliographystyle{aaai}
\bibliography{reference}
\end{document}